\documentclass{isprs} 
\usepackage{subfigure}
\usepackage{setspace}
\usepackage{geometry} 
\usepackage{epstopdf}
\usepackage[labelsep=period]{caption}  
\usepackage[british]{babel} 
\usepackage[hang]{footmisc}
\usepackage{amsmath}
\usepackage[hyphens]{url}
\usepackage{hyperref}
\usepackage{booktabs}
\usepackage{makecell}
\usepackage{multirow} 
\usepackage{amssymb}
\usepackage{svg}
\usepackage{natbib}

\makeatletter
\renewcommand\hyper@natlinkbreak[2]{#1}
\makeatother
\hypersetup{
    colorlinks=true,    
    citecolor=[rgb]{0,0,.6},
    urlcolor =[rgb]{0,0,.6},
    linkcolor = blue
    }

\begin{document}

\title{Multi-temporal crack segmentation in concrete structures using deep learning approaches}
\date{}


\author{
Said Harb\thanks{Corresponding author} , Pedro Achanccaray, Mehdi Maboudi, Markus Gerke}

\address{
Institute of Geodesy and Photogrammetry, Technische Universit\"at Braunschweig, Germany \\{(s.harb, p.diaz, m.maboudi, m.gerke)}@tu-braunschweig.de\\}
%



\abstract{
Cracks are among the earliest indicators of deterioration in concrete structures. Early automatic detection of these cracks can significantly extend the lifespan of critical infrastructures, such as bridges, buildings, and tunnels, while simultaneously reducing maintenance costs and facilitating efficient structural health monitoring. This study investigates whether leveraging multi-temporal data for crack segmentation can enhance segmentation quality. Therefore, we compare a Swin UNETR trained on multi-temporal data with a U-Net trained on mono-temporal data to assess the effect of temporal information compared with conventional single-epoch approaches. To this end, a multi-temporal dataset comprising 1356 images, each with 32 sequential crack propagation images, was created. After training the models, experiments were conducted to analyze their generalization ability, temporal consistency, and segmentation quality. The multi-temporal approach consistently outperformed its mono-temporal counterpart, achieving an IoU of $82.72\%$ and a F1-score of $90.54\%$, representing a significant improvement over the mono-temporal model's IoU of $76.69\%$ and F1-score of $86.18\%$, despite requiring only half of the trainable parameters. 
The multi-temporal model also displayed a more consistent segmentation quality, with reduced noise and fewer errors. These results suggest that temporal information significantly improves the performance of segmentation models, offering a promising solution for improved crack identification and long-term monitoring of concrete structures, even with limited sequential data. The dataset is available at: \url{https://github.com/saidharb/Multi-Temporal-Crack-Segmentation-Dataset.git}
}


\maketitle


\section{Introduction}

Infrastructures such as roads, bridges, dams, harbors, and buildings are essential to the sustainability and efficiency of economic activities, to ensure public safety, and to facilitate social interactions \citep{zhou-2023,maboudi2024Concerto}. Many of these infrastructures are constructed from materials such as concrete, asphalt, or stone and face various challenges, including fatigue stress, cyclic loading, and cumulative effects of time, along with increasing human and environmental pressure \citep{mohan-2018,konig-2022,kheradmandi-2022}. These factors can compromise structural integrity \citep{hamishebahar-2022} and diminish the asset value of these constructions \citep{kheradmandi-2022}. Consequently, there is a pressing need for effective Structural Health Monitoring (SHM) to maintain and prolong the lifespan of critical structures \citep{hamishebahar-2022}. This helps reduce maintenance and repair costs \citep{kheradmandi-2022,konig-2022} and prevents damage to humans and the environment \citep{mohan-2018}.

Numerous structures worldwide are in poor condition and require continuous monitoring and damage detection systems. For instance, German seaports feature approximately $3000\,\text{km}$ quay walls and $2500\,\text{km}$ facilities along the federal waterways. Throu\-gh\-out $140\,\text{km}$ of these quay walls, approximately $279$ million tons of goods are handled, underscoring the significant role of these structures in the German economy. However, the current state of quay walls and facilities is a concern. According to the Federal Waterways Engineering and Research Institute (BAW), $70\%$ of quay walls and facilities on waterways are in adequate condition. This situation foreshadows the considerable maintenance efforts required to ensure continued transportation of essential goods on German waterways \citep{alamouri-2024}. 

Looking across the Atlantic, a 2022 report highlighted that the United States has $2000$ dams and over $46\,000$ bridges with structural deficits \citep{konig-2022}, and $11\%$ of roads are classified as in poor or mediocre conditions \citep{zhou-2023}. Similarly, in the United Kingdom, approximately $10\%$ of roads are disrepaired \citep{konig-2022}. 
These examples illustrate the significant global need for infrastructure assessment and maintenance.


One of the first and most common indicators of structural degradation and reduced structural integrity is the presence of cracks on the surfaces of structures \citep{konig-2022,hamishebahar-2022}. Cracks begin at the microscopic level and continuously reduce the local stiffness of materials and create material discontinuities \citep{mohan-2018}. If left untreated, these cracks grow in size, and the cost and effort required to repair them increase accordingly. To mitigate consequential safety issues, it is crucial for SHM to accurately assess the state of a structure and identify indicators of future damage, such as cracks \cite{konig-2022}.


\sloppy
Traditionally, manual visual assessment by an inspector has been the standard method for identifying cracks. However, new machine learning (ML) techniques, particularly deep learning (DL), have begun to change this landscape. Manual assessments are costly, labor-intensive, time-consuming, and require highly trained experts. Despite this training, human factors can lead to subjective results \citep{hamishebahar-2022,mohan-2018,isprs-ABCInspekt-2021}. For instance, inspectors may experience fatigue during assessments, exhibit inconsistencies in their evaluations, or receive inadequate training \citep{konig-2022}. Moreover, the manual inspection can pose hazards to inspectors owing to unsafe structures \citep{kheradmandi-2022}. In some instances, the locations requiring inspection may be inaccessible to humans \citep{konig-2022}, or the inspection process could result in downtime for the structure \citep{konig-2022,AbcInspekt2024}, leading to interruptions in traffic on roads, bridges, or tunnels \citep{kheradmandi-2022}.

Given these disadvantages, automatic crack detection methods are required. Computer vision (CV) techniques, in conjunction with ML/DL, have proven to be effective for this purpose. In this subfield, image-based approaches have emerged as the most cost-effective methods because of the widespread availability of cameras \citep{hamishebahar-2022}. When mounted on UAVs, these systems can reach otherwise inaccessible locations without interrupting structure use \citep{mohan-2018}.


To the best of the authors' knowledge, and as stated by König et al. (\citeyear{konig-2022}), there is a lack of research regarding the use of multi-temporal data for CV tasks related to crack detection. In this context, we focus on the semantic segmentation of cracks using multi-temporal data, which is compared to approaches that utilize mono-temporal data. We aim to address the research gap in multi-temporal crack propagation data for semantic segmentation by providing a multi-temporal dataset and a corresponding deserialized mono-temporal dataset, which provide a foundation for future research. The contributions of this study can be summarized as follows:
\begin{itemize}
    \item Development and assessment of mono- and multi-temporal models for crack segmentation.
    \item Creation of a multi-temporal dataset and a corresponding deserialized mono-temporal dataset.
\end{itemize}

\section{Related works}

Crack detection techniques can be based on RGB, infrared, ultrasonic, laser, and other types of images \citep{mohan-2018}. In this study, we focused on camera-based detection, similar to the dataset used in our experiments. Many review papers categorize automatic crack detection methods into two fundamental types: rule- and data-driven approaches. Rule-driven methods rely on edge information of cracks, morphological operations, and thresholding techniques. These methods may struggle to perform adequately on less structured and more complex data because of their rigidity and inability to adapt to different contexts \citep{kheradmandi-2022,gong-2024,hamishebahar-2022}. By contrast, data-driven approaches outperform rule-driven approaches in managing complex scenes by learning useful feature representations from training data, particularly DL-based methods. As a result, they tend to be more accurate, easier to automate \citep{gong-2024} and computationally more efficient. Consequently, this study focused on exploring DL approaches.

\begin{figure}
    \centering
    \includegraphics[width=\linewidth]{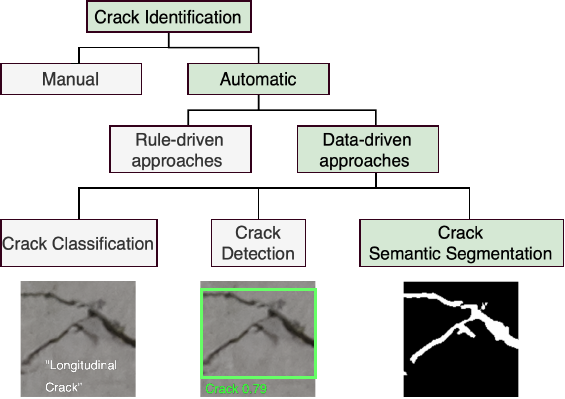}
    \caption{Overview of methods to analyze cracks.}
    \label{fig:categorization}
\end{figure}

\textbf{Semantic Segmentation of Cracks:} DL approaches can be applied to various tasks related to crack analysis, with the most significant being: crack classification (presence of cracks in the image, e.g., transverse or longitudinal cracks), crack segmentation (distinguishing between crack and background pixels), and crack detection (localization of cracks). An overview of this can be seen in Figure \ref{fig:categorization}. Semantic segmentation can be extended to instance segmentation, which aims to classify and localize each pixel of an individual crack \citep{konig-2022}. Most studies have emphasized crack segmentation because it provides the most precise localization and extent of cracks at the pixel level, making it particularly valuable for SHM \citep{hamishebahar-2022,gong-2024,zhou-2023}.

Semantic segmentation methods can be divided into two main categories: hybrid and pure segmentation. The former integrates classification or detection techniques with image processing methods, whereas the latter focuses solely on segmentation without these additional components. Given the scope of this study, we focus on the latter. According to Hamishebahar et al. (\citeyear{hamishebahar-2022}), pure segmentation architectures can be further divided into two fundamental types: encoder-decoder and non-encoder-decoder architectures. 
The encoder-decoder structure is a prominent paradigm in segmentation because it allows the use of various backbones to extract features \citep{gong-2024}. This architecture emphasizes the extraction of contextual information and multi-scale features, rather than the depth of the network. Typically, this is achieved by concatenating features from feature maps of different resolutions, thereby enabling the combination of high-level contextual information with low-level localization features \citep{gong-2024}. The encoder module is then responsible for extracting features from the input image while downsampling it by strided convolutions or attention mechanisms. The decoder module concatenates the information from the encoder and upsamples the extracted features to generate pixel-level predictions \citep{zhou-2023}. The encoder can be substituted by any model architecture that produces hierarchical feature maps. The most commonly used backbones, as noted by Gong et al. (\citeyear{gong-2024}) and Hamishebahar et al. (\citeyear{hamishebahar-2022}), include VGG \citep{simonyan-2015}, ResNet \citep{he-2016}, and transformers \citep{vaswani-2017}.

\textbf{Multi-temporal data:} The limited availability of multi-temporal crack propagation datasets has resulted in most crack segmentation research focusing on mono-temporal images (using datasets such as \textit{CrackTree} \citep{CrackTree}, \textit{CFD} \citep{CFD}, \textit{CRKWH100} \citep{CRKWH100}, \textit{DeepCrack} \citep{DeepCrack}, and \textit{Crack500} \citep{Crack500}
, ignoring the crack propagation effect over time. On the other hand, remote sensing (RS) exploits spatio-temporal features due to its access to satellite data from different locations around the world at different times.
In this context, RS approaches can be effectively extended to crack detection, as both fields utilize similar data types, namely high-resolution images. 

Numerous methods have been utilized in DL models for crack segmentation, with the most notable being those that rely on Convolutional Neural Networks (CNN) and the attention mechanism \citep{gong-2024,kheradmandi-2022}. In the first category, a distinction can be made between 1D (along the time dimension), 2D (along the spatial dimensions), and 3D (along the spatial and time dimensions) convolutions, depending on the type of relationships learned by the model \citep{zhong-2019}. The attention mechanism can function as the sole mechanism within a transformer architecture (as a feature extractor) or can be integrated into encoder-decoder architectures alongside CNN-based networks (using spatial or channel attention). This mechanism processes data from a global perspective and assigns weights based on the learned significance of information. Although the attention mechanism offers advantages over CNNs in capturing long-range dependencies, it may exhibit limitations in localization capabilities because of its large receptive field \citep{gong-2024,zhong-2019}.

\textbf{Mono- and Multi-temporal models:} One of the first models for semantic segmentation was the Fully Convolutional Network (FCN) \citep{shelhamer-2016}, which first reduces the size of the input image through convolution and pooling layers (encoder) and then upsamples the features again through deconvolutional layers (decoder) to generate a segmentation map for all pixels. Similarly, SegNet employs an encoder-decoder structure, where pooling indices from the encoder are used during upsampling in the decoder. Later, U-Net \citep{ronneberger-2015} further developed this structure by adding skip connections between the encoder and decoder to preserve spatial information, being successfully applied for crack segmentation \citep{konig-2022}.

Çiçek et al. (\citeyear{cicek-2016}) introduced a 3D U-Net, where 3D convolutional layers are employed to segment volumetric images. This model can be easily extended to multi-temporal data, where the third dimension is interpreted not as depth, but as the temporal domain. Xu et al. (\citeyear{xu-2019}) presented a Long Short-Term Memory (LSTM) multi-modal U-Net, which is a U-Net with multiple encoder paths designed to process multi-modal data for the segmentation of brain tumor data. An attempt was made to increase the learning capabilities of the model by not fusing the feature extraction of different channels. The decoder concatenates information from every encoder path and produces a segmentation mask. This multi-modal U-Net is applied to different slices of volumetric multi-modal input data. The output segmentation masks create a sequence of segmentation masks based on the depth of the image. This sequence is sequentially input into Convolutional LSTM (ConvLSTM) networks to capture the sequence dependencies. Therefore, volumetric data is fully segmented. Rustowicz et al. (\citeyear{rustowicz-2019}) employed a 2D U-Net in combination with a ConvLSTM for crop type segmentation using multi-modal and multi-temporal satellite data. The model has a U-Net type structure, where the encoder first extracts the features of each channel independently using CNNs, and these features are each fed into a ConvLSTM. Zhou et al. (\citeyear{zhou-2023}) proposed the Swin UNETR model, which also employs a U-Net architecture with a Swin Transformer as a backbone in the encoder, enhanced to process volumetric data. Feature extraction of the backbone relies solely on the attention mechanism, whereas CNNs are only employed within residual blocks in skip connections to refine the features. The decoder concatenates features from the feature maps with different resolutions and upsamples them to the size of the input data. Therefore, volumetric or multi-temporal data can be fully segmented. We selected this model for multi-temporal crack segmentation because it uses attention mechanisms to extract features and segment data at the original input size. To the best of the authors' knowledge, attention mechanisms have not yet been applied to multi-temporal crack segmentation.

\section{Materials and Methods}
To address the aforementioned research question, a multi-temporal dataset was created, and a Swin UNETR model was selected to serve as the foundation for a comparative analysis of mono- and multi-temporal approaches.

\subsection{Dataset}

\begin{figure}[t]
    \centering
    \includegraphics[width=\linewidth]{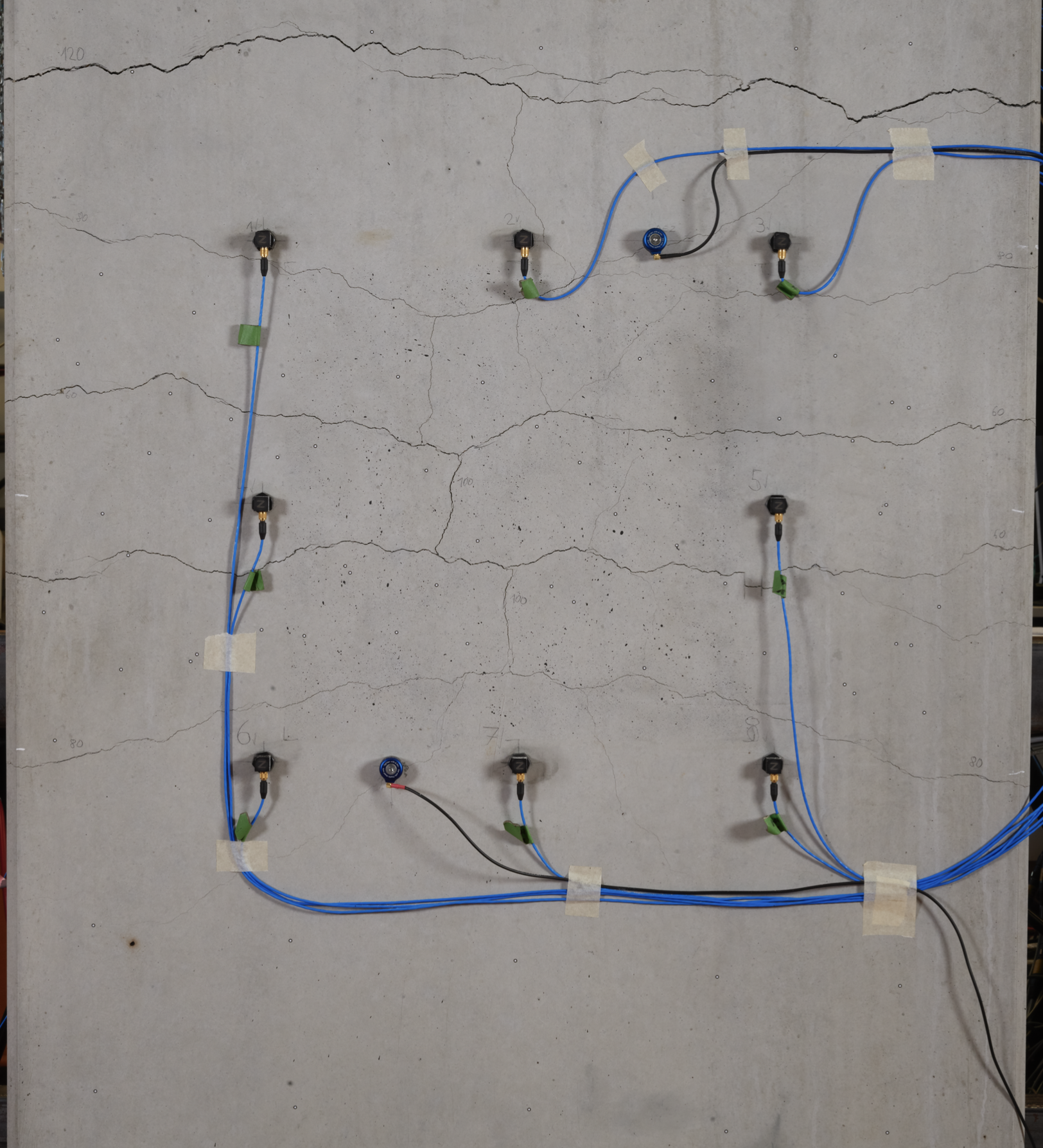}  
    \caption{Concrete block in the last epoch. This is the last stage of crack propagation on the concrete block. The preceding 24 images show the incremental crack propagation through time.}
    \label{fig:last_epoch}
\end{figure}

A temporal sequence of images depicting crack propagation is required to create a multi-temporal dataset for crack segmentation. For this purpose, a setup was constructed involving a vertical concrete slab with an area of $2\,\text{m}^2$ and thickness of $15\,\text{cm}$ (see Figure \ref{fig:last_epoch}), which was reinforced and positioned upright. A force was applied to the center of the slab, with the load incrementally increasing by $20\,\text{kN}$ at each load stage, referred to as epochs, until the maximum load was reached. Eight epochs were recorded before the concrete slab failed. During each epoch, high-resolution RGB images, aligned by reference points on the concrete slab, were captured using a mounted camera positioned at a specified distance \citep{backhaus-2024}. This study used $25$ images with a resolution of $11664 \times 8750$ pixels and a ground sampling distance of $0.3\,\text{mm}$ from the final epoch to generate training data. These images are in temporal order and provide a detailed view of the entire concrete block, with the progression of cracks between the images visible upon visual inspection. 

\begin{figure*}[t]
    \centering
    \includegraphics[width=0.99 \textwidth]{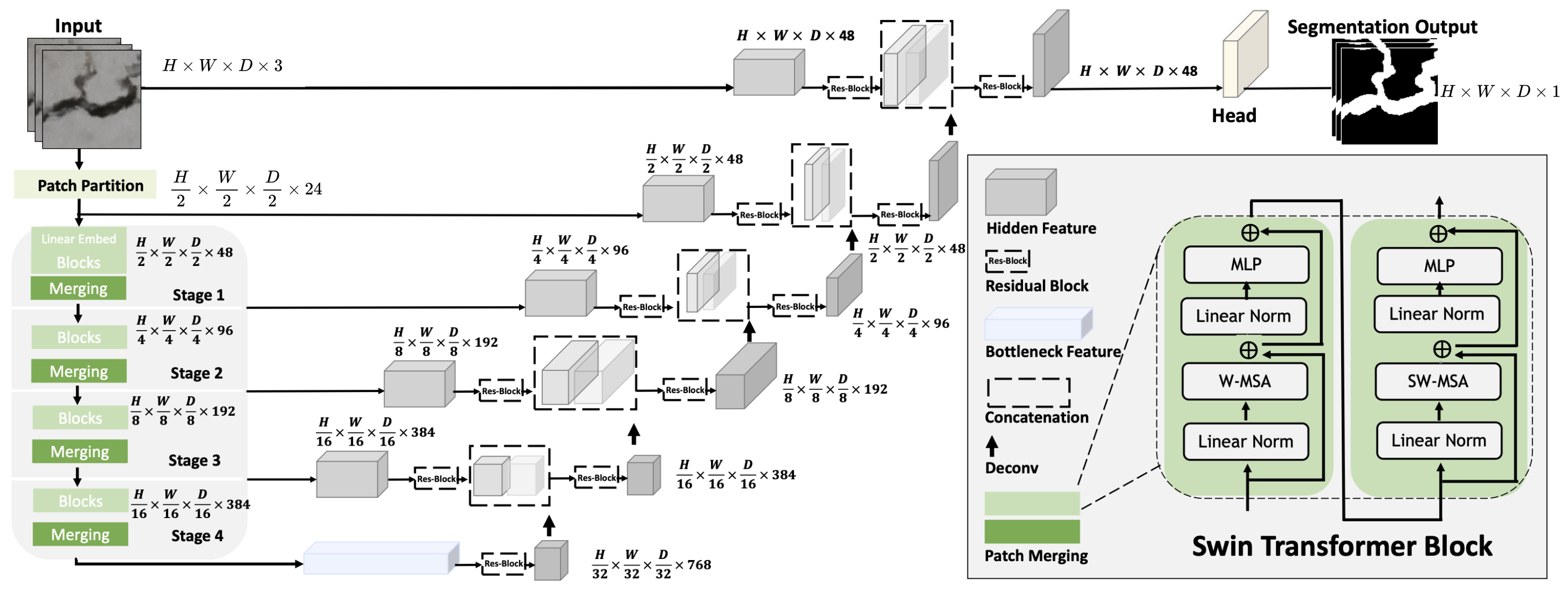}
    \caption{Swin UNETR architecture. A 3D Swin Transformer is used as a feature extractor, and in the decoder, the feature maps are concatenated and upsampled to the original input size. Adopted from \citep{hatamizadeh-2022}.}
    \label{fig:swinUNETR_architecture}
\end{figure*}

The creation of mono- and multi-temporal datasets can be divided into three phases: labeling, post-processing, and aggregation. In the labeling phase, because manual annotation of all images is tedious and extremely time-consuming, a set of three images was manually annotated and used to train a segmentation model (auxiliary U-Net) to assist the annotation process of the remaining 25 images. The post-processing phase involves the removal of false positives (pencil drawings, concrete pores, shadows, and cables) and segmentation errors (isolated groups of pixels smaller than five pixels) in the auxiliary U-Net predictions. Then, morphological operations were applied (closing with a $3\times3$ kernel), followed by manual cleaning to refine the crack annotations. Finally, the aggregation phase consists of fulfilling the model requirements for training, as the multi-temporal model requires sequences of 32 images (cf. Section \ref{ssec:Methodology}), but only 25 are available. To address this, seven images at evenly spaced intervals were duplicated to expand the sequence to 32 images.

\begin{table}[htp]
    \centering
    \setlength{\tabcolsep}{4pt}
    \renewcommand{\arraystretch}{1.1}
    \begin{tabular}{c lrcc}
        \toprule
        \makecell{\textbf{Type}} & & \makecell{\textbf{Sample}\\\textbf{size}} & \makecell{\textbf{Crack image}\\\textbf{ratio (\%)}} & \makecell{\textbf{Crack pixel}\\\textbf{ratio (\%)}} \\
        \midrule
        \multirow{4}{*}{\makecell{\textbf{Multi-}\\\textbf{temp.}}} 
        & All     & $1\,356$    & $66.7$ & $1.2$ \\
        & Train   & $813$       & $66.5$ & $1.2$ \\
        & Val     & $271$       & $66.8$ & $1.3$ \\
        & Test    & $272$       & $66.9$ & $1.2$ \\
        \midrule
        \multirow{4}{*}{\makecell{\textbf{Mono-}\\\textbf{temp.}}} 
        & All     & $43\,392$   & $40.1$ & $1.2$ \\
        & Train   & $26\,016$   & $40.1$ & $1.2$ \\
        & Val     & $8\,672$    & $39.1$ & $1.3$ \\
        & Test    & $8\,704$    & $41.3$ & $1.2$ \\
        \bottomrule
    \end{tabular}
    \caption{Statistics for multi-temporal and mono-temporal datasets.}
    \label{tab:dataset_statistics}
\end{table}

\sloppy
To create the multi-temporal dataset, images and corresponding segmentation targets were stacked in temporal order. From these stacks, non-overlapping spatial patches of size 
$128\times128$ pixels were extracted, resulting in a total of 5,632 samples, each consisting of a sequence of 32 consecutive crack propagation image-target pairs. As expected, there were several samples without cracks. To balance the dataset, all multi-temporal samples with cracks were included, and samples without cracks were randomly selected and removed to achieve a 2:1 ratio of images with cracks to those without cracks. The training, validation, and test sets were created at the following proportions: 60\%, 20\%, and 20\%, respectively. To generate the mono-temporal dataset and ensure that both models have access to exactly the same data, the multi-temporal samples within the splits were deserialized. Table \ref{tab:dataset_statistics} shows the dataset statistics.

The sample size of the multi-temporal dataset is $32$ times smaller than in the mono-temporal dataset because of sequential stacking. The deserialization of the multi-temporal dataset leads to a decreased crack image ratio in the mono-temporal dataset.
        
        
        

\subsection{Methodology}
\label{ssec:Methodology}
The methodology employed in this study is as follows. First, the datasets were created and annotated as previously mentioned. Second, a multi-temporal model was trained to perform the segmentation task on the datasets. Finally, the assessment of the multi-temporal approach will be quantitative, based on metrics, and qualitative, based on a visual inspection of the segmentation results. 

We selected the Swin UNETR model for multi-temporal crack segmentation to extract features from the spatial and temporal domains using its attention mechanisms.

\textbf{Swin UNETR:} The Shifting Window U-Net Transformer (Swin UNETR) was introduced by Hatamizadeh et al. (\citeyear{hatamizadeh-2022}) and extends the Shifting Window Transformer (Swin Transformer) \citep{liu-2021} to perform semantic segmentation on three-dimensional inputs. This is achieved by combining a Swin Transformer adapted for three-dimensional inputs with a U-Net architecture. In Figure \ref{fig:swinUNETR_architecture} the architecture of Swin UNETR is depicted. The input tokens are patches of pixels, which are embedded into a feature dimension $C$ in the input layer. The Swin Transformer then serves as a feature extractor in the encoder path, whereas the decoder path reassembles the original input shape and predicts each voxel using the feature maps produced by the Swin Transformer. The encoder path features five downsampling operations, resulting in a reduction of resolution by a factor of $2^5=32$. Therefore the minimum input resolution of samples is $32\times32\times32$ pixels. The Swin Transformer enables dense prediction tasks for the transformer architecture, which is not feasible in the preceding Vision Transformer (ViT) \citep{dosovitskiy-2021}, owing to its quadratic computational cost. 

        
        
        

The computational complexity of the Swin Transformer is reduced to a linear order by applying the attention mechanism only in local windows instead of the whole image and shifting the window configuration to introduce the global context in two consecutive Swin Transformer blocks. The attention mechanism is embedded within the window-masked self-attention and shifted-window masked self-attention modules in transformer encoder blocks, which were presented by Vaswani et al. (\citeyear{vaswani-2017}) in the original transformer. Patch-merging layers decrease the resolution of feature maps in each stage, thereby creating a hierarchical structure of feature maps that can be used for dense prediction tasks.

\textbf{U-Net:} The U-Net was introduced by Ronneberger et al. \citep{ronneberger-2015} and features a encoder-decoder structure employing convolutional down- and up-sampling layers as well as skip connections. Due to its simple architecture and robust performance it is often used in crack segmentation \citep{hamishebahar-2022, kheradmandi-2022} and will be considered as a baseline. To produce outputs of the same size as the inputs, padding was aplied in this work.

\section{Experimental Setup and Results}
In order to compare the mono- and multi-temporal approach this section will show the different experiments conducted and draw a conclusion based on the results of the experiments.

\subsection{Experimental Setup}
Three experiments were conducted by modifying the size and transformations of the training set. The first experiment examined the effect of a reduced sample size, where the mono-temporal training and validation datasets for the U-Net were matched to the smaller size of the multi-temporal dataset. The second and third experiments focused on data augmentation (DA) by applying five transformations (horizontal and vertical flips, brightness, contrast, and blurring) with a probability of $0.5$. The second experiment applied DA to expand the multi-temporal SwinUNETR dataset, whereas the third applied DA to the mono-temporal U-Net dataset.

As baselines, the following models were used: a mono-temporal U-Net (31M parameters) and a multi-temporal SwinUNETR (15.7M parameters), both implemented using PyTorch \citep{paszke-2019} and trained on a Tesla P100 GPU. We employed early stopping with a patience of 20 epochs, monitoring the validation Intersection over Union (IoU) for the crack class, along with a step learning rate scheduler that adjusted the learning rate by a factor of $0.1$ if the validation loss for the class crack does not decrease for $10$ epochs. For the batch size, a value of $4$ was chosen. Both models processed $128\times128$ images owing to memory constraints. SwinUNETR was taken from the Medical Open Network for AI (MONAI) \citep{cardoso-2022} framework and has an input channel size of $3$, an output channel size of $1$, and a feature size $C=24$; otherwise, the configuration is the default one described in the MONAI documentation.

\subsection{Results}

\subsubsection{Quantitative analysis:}
Table \ref{tab:results} summarizes the results obtained on the test set in the three experiments and the baseline models (BL) in terms of different quantitative metrics: IoU, Precision (P), Recall (R), and F1-score (F1). DS and DA are the downsampled dataset and dataset with data augmentation, respectively. We can see that the BL multi-temporal Swin UNETR outperformed all mono-temporal U-Net models. BL Swin UNETR achieved the highest metric scores, with an IoU of $82.7\%$ and an F1-score of $90.5\%$. Swin UNETR DA (experiment 2) performed only slightly worse than BL Swin UNETR, a difference that may be attributed to the inner randomness of each model, given that the metrics are closely aligned. Notably, the recall is higher in the DA model compared to the BL model. Overall, both models exhibited similarly strong performance on the test set, with balanced P and R, indicating robust segmentation capabilities.

        
        
        

\begin{table}[htbp]
    \setlength{\tabcolsep}{4pt}
    \centering
    \begin{tabular}{clc ccccc}
        \toprule
        \multirow{2}{*}{\textbf{Exp.}}                      &\multirow{2}{*}{\textbf{Model}}    
        & \multirow{2}{*}{\textbf{DS}}                & \multirow{2}{*}{\textbf{DA}} 
        & \multicolumn{4}{c}{\textbf{Metrics (\%)}}    \\
        
        \cmidrule(lr){5-8} 
        
        &&&&  \textbf{IoU}$\mathbf{\uparrow}$ & \textbf{P}$\mathbf{\uparrow}$ & \textbf{R}$\mathbf{\uparrow}$ & \textbf{F1}$\mathbf{\uparrow}$ \\
        \midrule
        
        BL    &U-Net                      &--- & ---& $76.7$ & $87.0$ & $86.6$ & $86.8$  \\
        BL    &\makecell[l]{Swin UNETR}  & ---& ---& $\mathbf{82.7}$ & $\mathbf{91.9}$ & $89.2$ & $\mathbf{90.5}$  \\
        $1$&U-Net & \checkmark &  ---                         & $71.9$ & $80.5$ & $87.1$ & $83.7$  \\
        $2$&\makecell[l]{Swin UNETR} & --- &\checkmark  & $82.5$ & $91.2$ & $\mathbf{89.6}$ & $90.4$ \\
        $3$&U-Net & --- &\checkmark            & $75.7$ & $86.0$ & $86.3$ & $86.2$  \\
        \bottomrule
    \end{tabular}
    \caption{Results overview. Metrics obtained by the selected methods in the test set. BL: baseline, DS: downsampled dataset, DA: dataset with data augmentation.}
    \label{tab:results}
\end{table}

The best-performing U-Net on the test set was the BL U-Net model, which achieved an IoU of $76.7\%$ and an F1-score of $86.2\%$. A similar trend is observed in the Swin UNETR models, where the model trained with DA performs marginally worse than the BL model on the original dataset. Notably, the F1-score of the BL model was only $0.63$ percentage points higher than that of the DA U-Net from experiment 3. In contrast, the U-Net from experiment 1, which was trained on a downsampled (DS) set (denoted as DS U-Net), showed a larger performance gap compared to both the BL model and the DA U-Net. The IoU and F1-score for the DS U-Net are both more than $3$ percentage points lower than those of the DA U-Net and BL models. Furthermore, P and R were balanced for all U-Net experiments, except for experiment 1. In experiment 1, R was the highest among all U-Net experiments on the original dataset; however, this model exhibited a lower P compared to P across all other U-Net experiments. In summary, the results indicated that the multi-temporal approach consistently outperformed the mono-temporal approach across the test set. This suggests that the multi-temporal model effectively learned the temporal dependencies between sequential images, leading to improved segmentation results. 


\subsubsection{Qualitative analysis:}
In addition to the metrics, it is crucial to visually inspect the predictions of both the mono- and multi-temporal models to assess whether the multi-temporal model successfully learns temporal dependencies, in contrast to the mono-temporal model. For this purpose, the predictions of the baseline models were used.

\begin{figure}[t]
    \centering
    \includegraphics[width=1.0\linewidth]{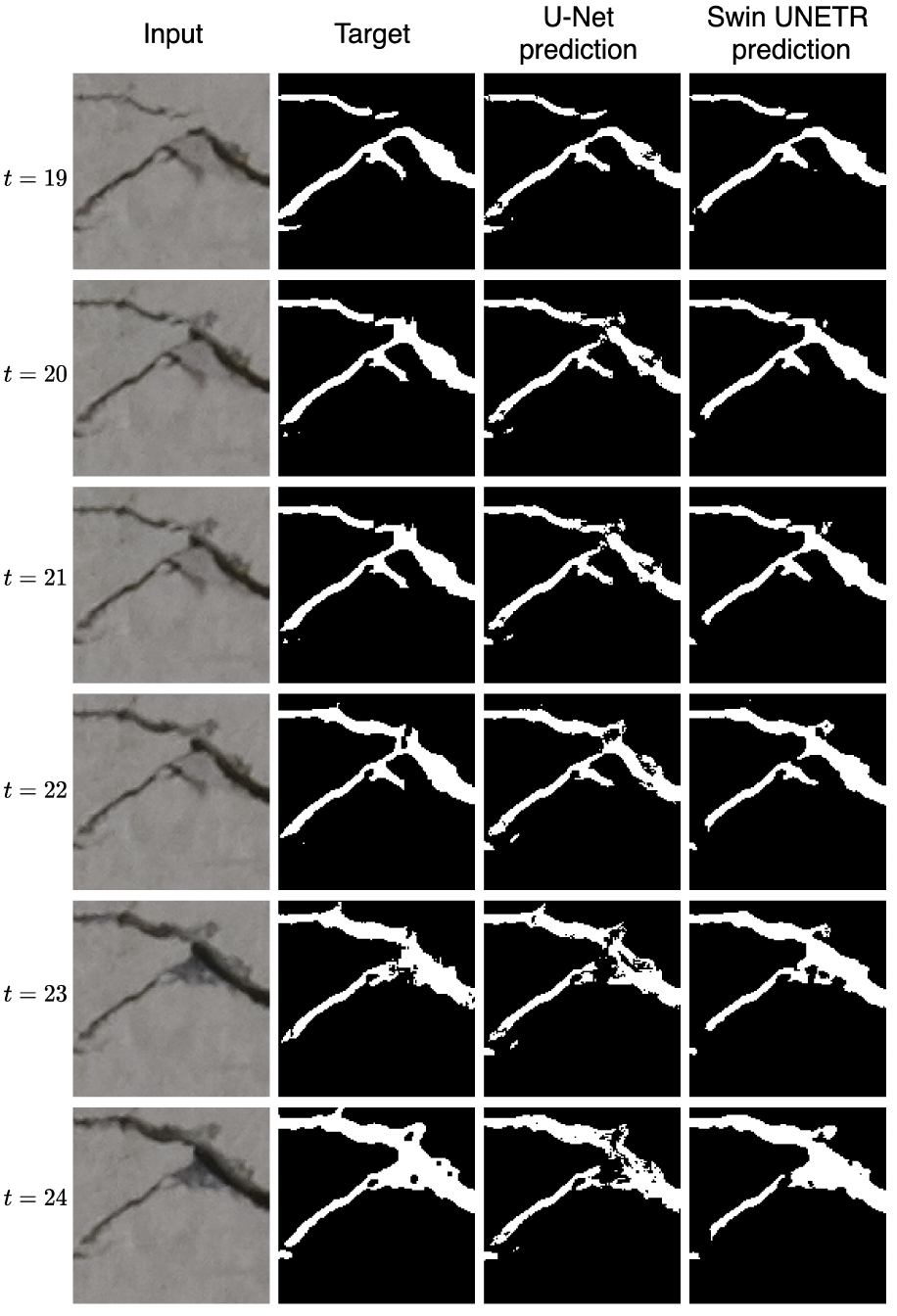}
    \caption{Comparison of U-Net and Swin UNETR sequential predictions. The Swin UNETR produces smoother and more consistent predictions containing less noise than the U-Net.}
    \label{fig:246}
\end{figure}

A brief visual inspection of numerous examples revealed that the multi-temporal Swin UNETR exhibits superior segmentation quality, aligned with the expectations set by the metrics. This improved performance was observed in several aspects. First, the predictions made by the Swin UNETR were smoother and contained less noise than those of the U-Net. This suggests that the Swin UNETR  effectively learned the temporal dependencies associated with crack propagation, resulting in greater confidence when consistently classifying the same region of an image as a crack. Second, although U-Net captures the general shape of the cracks, it introduces a substantial number of artifacts into its predictions. This inconsistency is particularly evident in Figure \ref{fig:246}, where Swin UNETR demonstrates greater consistency in predicting the crack shape and accurately classifying regions according to crack propagation. In contrast, U-Net's predictions vary significantly from image to image, primarily because of the noisy outputs. Consequently, more segmentation holes appear in the U-Net predictions, which are significantly less prevalent in those from Swin UNETR. While U-Net generally captures the edges of cracks well, it often lacks the same level of detection quality at the center of the cracks. 

\begin{figure}[t]
    \centering
    \includegraphics[width=\linewidth]{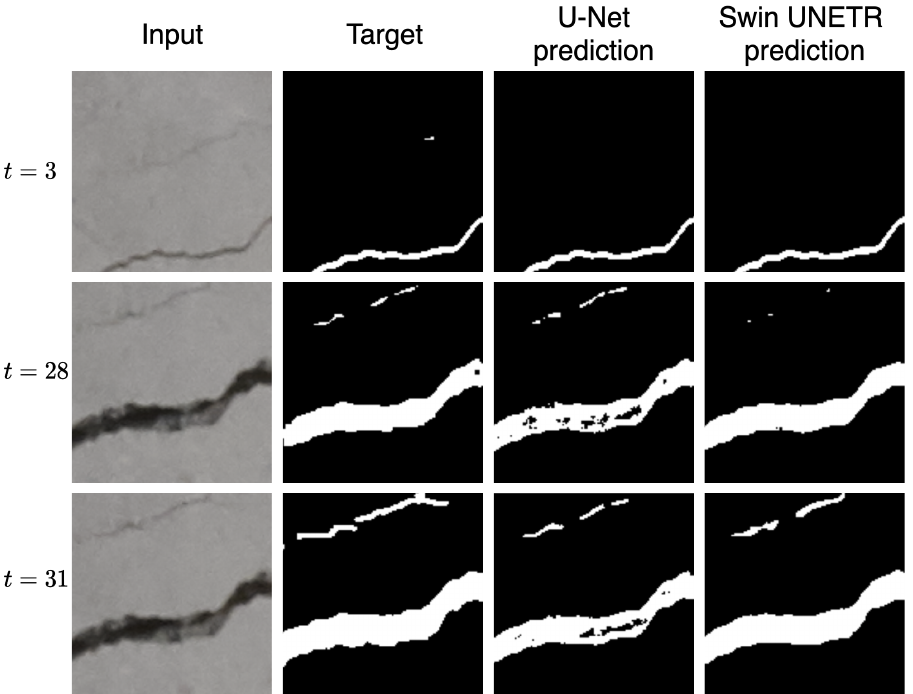}
    \caption{Early stage vs. late stage crack segmentation. Both models perform well on early stages cracks except for thin and low contrast cracks, which are also seldom in the targets. Only the Swin UNETR manages to maintain adequate segmentation quality in later stages of crack propagation.}
    \label{fig:83}
\end{figure}

U-Net and Swin UNETR demonstrated similar performance on early stage cracks, as illustrated in Figure \ref{fig:83}, where both models produce consistent predictions that accurately depict the target. However, thin and low contrast cracks are a challenge for both models. In the later stages of crack propagation, as shown in the subsequent images of Figure \ref{fig:83}, U-Net's predictions become less consistent and exhibit numerous holes not present in the target. In contrast, the Swin UNETR maintains a high level of consistency with no holes in its segmentation.



\subsubsection{Evaluation of false positives and false negatives: }
On the concrete slab, there were several distinct challenges for both models to handle, such as pencil-drawn numbers, sensors, cables, thin and low-contrast cracks, and cavities which all proved to be potential sources for false positive classifications. Generating predictions for the entire concrete slab reveals whether the models learned to handle these challenges.

\begin{figure}[t]
    \centering
    \includegraphics[width=\linewidth]{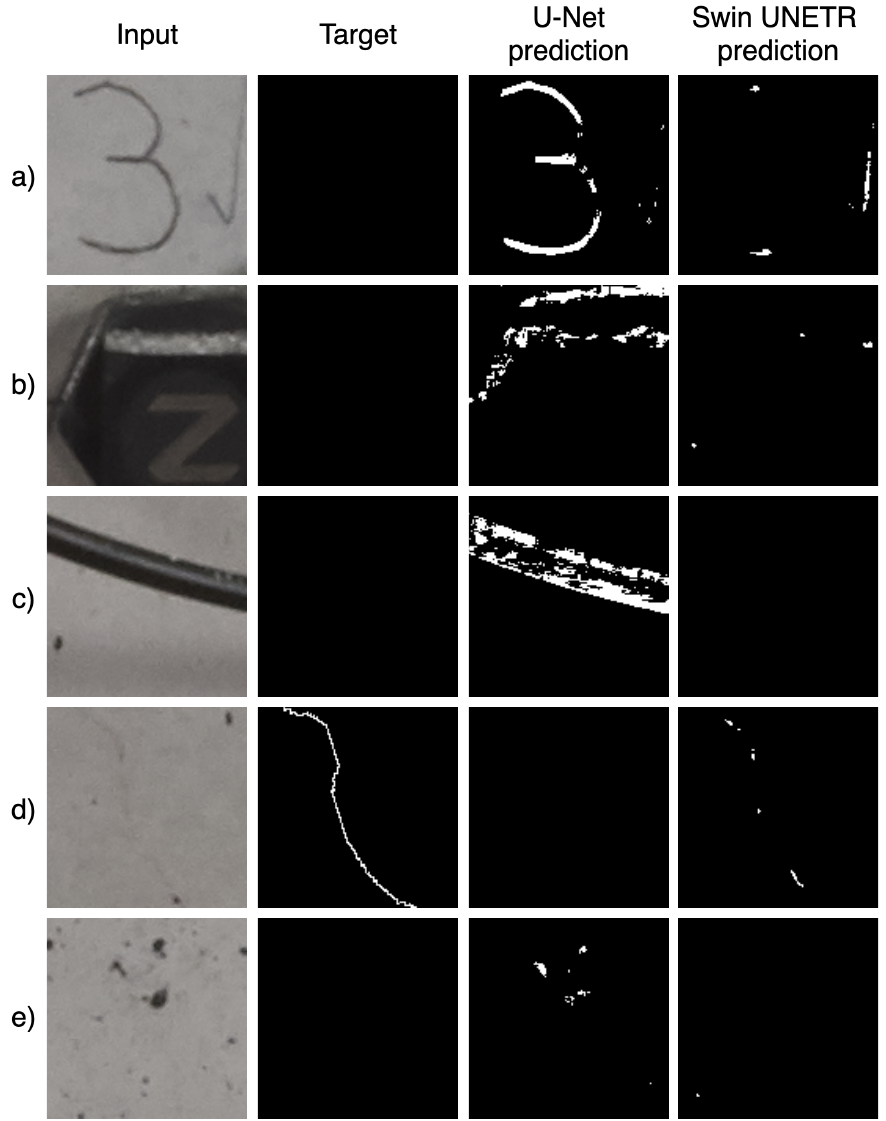} 
    \caption[Comparison of segmentation errors]{Comparison of segmentation errors. a) Pencil drawings, b) Sensor, c) Cable, d) Thin and low contrast crack, e) Cavities}
    \label{fig:error_comparison} 
\end{figure}

The most significant segmentation error is the labeling of the numbers drawn on the concrete with a pencil as cracks. Visual inspection of the input images revealed $18$ visible numbers in a single image. By examining the segmentation masks of both models on the complete images, it is possible to count the numbers labeled as cracks by each model, providing insight into whether the models can differentiate numbers from cracks. In the final segmentation mask of the Swin UNETR, nine numbers are visible, whereas U-Net identifies $17$ numbers as cracks in the figure. This significant difference underscores the interpretation that the multi-temporal Swin UNETR has learned that these numbers do not possess typical crack features such as propagation over time, and are therefore not cracks. In contrast, U-Net relies more on less sophisticated features, such as the shape and color of the numbers, to determine if they represent cracks. Some examples of these challenges are presented in Figure \ref{fig:error_comparison}. The pencil-drawn number is completely recognized as a crack by U-Net, as well as parts of the sensor and cable. On the other hand, Swin UNETR managed to overcome these challenges; in particular, even thin and low-contrast cracks were partly segmented. Cavities in the concrete proved to be a source of noise in the segmentation masks, as both models classified some cavities as cracks. However, U-Net does this to a larger extent than Swin UNETR.


Next, there are $10$ sensors with cables attached to the concrete, which present segmentation challenges for the models. Both models often misclassified the edges of these sensors and cables as cracks. In examining the segmentation masks, U-Net misclassified the edges of nine out of ten sensors as cracks, whereas Swin UNETR only did this for one sensor. The dark color of the sensors contributes to this issue because their edges are sometimes mistaken for cracks. The cables attached to the sensors pose an additional challenge to both models. Their long, thin, and darker appearance makes it difficult to distinguish them from the actual cracks. In this case, Swin UNETR outperformed U-Net. U-Net incorrectly classified most of the darker cables as cracks, resulting in a significant false positive. In contrast, the Swin UNETR did not cause this error. However, it misclassifies the edges of the blue cables as long, thin cracks. The final prominent error observed is the misclassification of cavities, which appear as small dark holes in concrete. A comparison of the segmentation masks reveals that Swin UNETR has significantly fewer misclassified cavities than U-Net, which is reflected in the higher IoU of Swin UNETR.

False negatives in this context are cracks which were not identified by the models. The Swin UNETR has a higher recall than the U-Net and therefore is able to find more cracks, like in Figure \ref{fig:error_comparison} d). However, it is important to note that the performance of both models for thin and low-contrast cracks is lower than that for more common cracks.

\section{Conclusions}
This study examined the effect of multi-temporal data on the performance of semantic segmentation for crack images. The Swin UNETR was trained on a multi-temporal crack dataset, whereas a U-Net was trained on the deserialized version of the same dataset for comparison. Swin UNETR achieved an IoU of $82.72\%$ and an F1-score of $90.52\%$, whereas U-Net resulted in an IoU of $76.69\%$ and an F1-score of $86.81\%$ on the test set. This indicates that the multi-temporal approach outperformed the mono-temporal approach, even when utilizing only half the parameters compared to U-Net.


The multi-temporal Swin UNETR demonstrated superior segmentation quality compared with the mono-temporal U-Net. Swin UNETR predictions exhibited greater consistency, reduced noise, and enhanced ability to distinguish cracks from visually similar features in the images. By leveraging the entire time-series data, the multi-temporal model significantly improved its performance in ambiguous image regions, resulting in more confident predictions. These properties enable a use-case of multi-temporal segmentation models in SHM in long-term scenarios. On one hand such models have less false negatives, thereby triggering fewer false alarms and on the other hand the superior segmentation quality facilitates a better detection of actual cracks. Due to available temporal information, small and continuous misalignments of images will be learned by the model, making it easier for deployment in real-life scenarios. By employing data augmentation during training, models can be adapted to be robust against changes to the surface due to weathering. 

In contrast, mono-temporal U-Net encountered several challenges. It struggled to segment larger cracks without introducing artifacts, and often produced noisy predictions. Furthermore, U-Net frequently failed to differentiate between actual cracks and visually similar features, highlighting its limitations in this specific application. Access to temporal information proved to be a crucial advantage for the Swin UNETR, enabling it to make informed decisions based on the evolution of features over time. This temporal context allowed the model to better interpret ambiguous areas and reduce false positives, resulting in more accurate and reliable crack segmentation.


\textbf{Acknowledgment}:
The authors would like to thank the DFG (German Research Foundation) for the support within the project “Optical 3D-Bridge-Inspect” – Project Number 501682769 as part of DFG’s Priority Programme 2388 “Hundred plus”. This research was supported by DFG (German Science Foundation) through a Major Research Instrumentation, project number 461109100.

{
	\begin{spacing}{1.17}
		\normalsize
		\bibliography{ISPRSguidelines_authors} 
	\end{spacing}
}

\end{document}